\documentclass[conference]{IEEEtran}
\IEEEoverridecommandlockouts
\usepackage{cite}
\usepackage{amsmath,amssymb,amsfonts}
\usepackage{graphicx}
\usepackage{textcomp}

\usepackage{tabularx}
\usepackage{xcolor}
\usepackage{color}
\usepackage{booktabs}

\usepackage{colortbl}
\definecolor{mycolor}{rgb}{0.9, 0.0, 0.0}
\usepackage{caption}
\usepackage{array}
\usepackage{dblfloatfix}
\usepackage{bigstrut}
\usepackage{multirow}
\usepackage{multicol}
\usepackage{enumitem}
\usepackage{url}

\usepackage{cite}
\usepackage{amsmath,amssymb,amsfonts}
\usepackage{algorithmic}
\usepackage{graphicx}
\usepackage{textcomp}
\usepackage{xcolor}
\def\BibTeX{{\rm B\kern-.05em{\sc i\kern-.025em b}\kern-.08em
    T\kern-.1667em\lower.7ex\hbox{E}\kern-.125emX}}
\begin{document}

\title{Robust Face Morphing Attack Detection Using Fusion of Multiple Features and Classification Techniques
}

\author{Jag Mohan Singh\textsuperscript{1} \quad \quad   Sushma Venkatesh\textsuperscript{2} \quad \quad  Raghavendra Ramachandra \\ \textsuperscript{1} 
\textsuperscript{1}Norwegian University of Science and Technology (NTNU), Norway.
\textsuperscript{2}AiBA AS, Norway. \\
\texttt{email:jag.m.singh@ntnu.no; sushma@aiba.ai;raghavendra.ramachandra@ntnu.no} \\
}

\maketitle

\begin{abstract}
The face morphing process will combine two or more facial images to generate a single morphed facial image demonstrating Face Recognition Systems (FRS) vulnerability. The attack potential of the morphing image directly depends on the perceptual image quality, and when generated with no visible artefacts, it can deceive both human observers and automatic FRS. The current softwares for face morphing generates a morphing image with ghosting artefacts, especially in the eye region, nose and mouth area, which may serve as a potential cue to detect morphing attacks. Hence in this work, we introduce a new dataset comprising 10710 facial images before and after manual post-processing to reduce the visual artefacts and to generate high-quality attacks. Further, we propose a novel single image-based Morph Attack Detection (S-MAD) technique based on the ensemble of features and classifiers using the scale-space domain. The novel concept in the proposed method is the multi-level fusion that combines the comparison scores from different features and classifiers.            
Extensive experiments are carried out on the newly generated high-quality face images with (i) Morphs before post-processing and (ii) Morphs after post-processing. Further, the experiments are also carried out on two different mediums such as (i) Digital and (ii) Print-scan (or re-digitized) with and without compression. Extensive experimental results are performed to benchmark the detection performance with the existing S-MAD techniques. Obtained results indicate the best performance of the proposed method over existing methods.
\end{abstract}

\section{Introduction}\label{sec:introduction}
Biometrics has been widely studied and applied globally for person identification \cite{JainFlynnRoss-HandbookOfBiometrics-Springer-2007}. The trustworthiness of biometric features has gained immense popularity over multi-factor authentication. Among several other physiological modalities like a fingerprint, palmprint, finger vein and iris, face biometrics-based applications have had a wide range of applications for several decades. The face is a unique modality and humans easily identify an individual based on facial features. As identification of a person based on facial features can be achieved through the naked eye, facial biometrics has been well accepted for national ID programs and security-related applications, especially in highly secure places such as border control scenarios.

Although Face Recognition Systems(FRS) are widely installed to provide reliable person identification and recognition, it also encounters threats due to various attacks that highlight the vulnerability of FRS. Presentation attacks, adversarial attacks, and imposter attacks are some example attacks that pose a risk to the reliable performance of FRS \cite{PADsurvey_2017, vakhshiteh2021adversarial}. In addition to these attacks, a face-morphing attack is one such attack that can efficiently make the FRS vulnerable, especially in border control applications. Although face morphing was initially performed merely for entertainment, it has gradually transformed into a potential threat in the recent past \cite{Ferrara-TheMagicPassport-IJCB-2014}. As face morphing is achieved by blending the facial features of two or more facial identities to generate a morphing image, this will lead to the vulnerability of FRS to reliably recognize the person.

Based on the International Civil Aviation Organisation (ICAO) recommendation, the face is the prominent modality employed for person recognition and verification in the border control scenario \cite{ICAO-9303-p12-2015, ICAO-9303-p9-2021}. Hence all passport holders must enroll their facial image in the eMRTD to serve as an identification document for border control authority during travel. Face enrolment procedure varies with the country's passport application procedure. Scandinavian countries have installed a photo booth to perform live capture of the facial image \cite{Kalvet2018LiveEF}. However, most Asian countries accept printed passport-size facial images during the application process \cite{india2019portal}. But New Zealand, Ireland and the UK have a web portal where the applicant has to upload the facial image for the passport renewal process \cite{nzvisa2019portal, ukpassport2019portal}. 
Even though the facial image undergoes manipulation and makes it easier to identify the existence of morphing, the availability of a variety of high-quality morphing software makes it challenging even for an expert human observer.

Several open-source morphing software yield superior quality morphed facial image that does not require any technical expertise \cite{FantaMorph,3Dmorph,Faceswap,MorphThing,Facemorpher}. Hence a person with malicious intentions can easily generate a morphed facial image with a look-alike accomplice's facial image and successfully submit for the passport enrolment process. As it is challenging to detect unknown facial identities from the morphing image, even a trained border control official finds it difficult to detect the existence of morphing \cite{RobinS-MADHuman-CognitiveResearch-2019, Robertson-FraudulentFaceMorphs-PLOS-ONE-2017}. Eventually, the morphing facial image will be enrolled in the eMRTD that can be claimed by both the identities involved in the morphing process. This disregards the rule of single ownership for the passport/eMRTD document and eventually creates a loophole in the security.
Considering the risk of face morphing and its impact on building a secure society, extensive research has been performed to generate robust techniques for Morph Attack Detection(MAD) \cite{Chaudhary_2021_CVPR, Raghavendra-MADusingScaleSpaceFeature-ISBA-2019, Venkatesh-SingleMAD-Fusion-2020, Ferrara-Demorphing-TIFS-2018, Ferrara-FaceDemorphing-IEEE-EUSIPCO-2018, Spreeuwers-MorphingEvaluation-EUSIPCO-2018,singh2019robust,singh2022reliable}. Based on the MAD techniques developed by several researchers, morph attack detection techniques can be broadly classified into single image-based MAD (without reference image) and differential image-based MAD (with reference image). S-MAD techniques are applicable where single facial image-based person verification is required. In the case of the passport renewal process in Ireland, \cite{PassportOnline}, since it is an online passport service, the applicant's facial image must be uploaded into the web portal. As no supervision exists while uploading the facial image into the web portal, an applicant with malicious intentions may end up uploading the morphed facial image.

Hence several researchers have investigated the problem of face morphing and developed reliable techniques. The first work on the S-MAD technique is investigated by Raghavendra et al. \cite{Raghavendra-DetectingMorphedFace-BTAS-2016} using the texture-based approach. Since then, several S-MAD approaches have been proposed that can be broadly divided into \cite{Sushma_MAD_Survey} three types (1) Hand-crafted features: These techniques include the different types of features such as: texture-based \cite{Makrushin-Dempster-ShaferTheory-IEEE-EUSIPCO-2019} \cite{Makrushin-BenfordLaw-2018} , time-frequency based \cite{aghdaie2021detection}, color based \cite{Raghavendra-MADusingScaleSpaceFeature-ISBA-2019}, residual noise \cite{Venkatesh-DeepColorMAD-IPTA-2019}, image quality based \cite{Hildebrandt-StirTrace-IWBF-2017, Seibold-Reflection-EUSIPCO-2018, Scherhag-PRNU-TBIOM-2019} (2) Deep learning features: These includes the use of pre-trained deep CNN networks \cite{Seibold-FaceMorph-IEEE-BIOSIG-2019, Raghavendra-DNNMorphingDetection-CVPR-2017, Matteo_PrintScan_2019}, fusion of pre-trained CNNs \cite{Venkatesh_2020_WACV, aghdaie2021detection}, pixel based DCNN MAD \cite{damer2021pw} (3) Hybrid Features: These MAD techniques are based on using multiple features and classifiers for face morphing detection. The outcome of the multiple classifiers is combined at either feature or comparison level. Several works proposed in this category includes \cite{aghdaie2021morph}, \cite{ohaire2021adversarially}, \cite{aghdaie2021detection}, \cite{Raghavendra-DetectingFaceMorphing-CVIP-2018}. Among these techniques, the hybrid approaches have indicated the best performances in detecting face-morphing attacks.      

All the available State-Of-The-Art (SOTA) techniques are evaluated on the morphed datasets that are not manually and professionally post-processed. Even though the early work \cite{Raja-SOTAMD-DatabaseEvaluationBenchmarking-TIFS-2020} attempts to use the manual post-processing morphs, the dataset size is tiny. In this work, we introduce a new dataset to benchmark the S-MAD techniques' performance systematically. The new dataset is constructed using different mediums such as: digital, print-scan using a DNP printer and print-scan using a Canon printer. We have used standard (Canon) and sublimation (DNP) printers to study the influence of printer noise on face morphing attack detection. The new dataset consists of a total of 10710 facial images before and after post-processing. Further, we have also proposed a new S-MAD technique based on the multi-level fusion of ensemble features and classifiers.  

To efficiently evaluate the performance of the proposed MAD technique and its performance over SOTA MAD techniques, we investigate the following research questions that facilitate this study.

\begin{figure}[htp]
	\centering
	\includegraphics[width=1\linewidth]{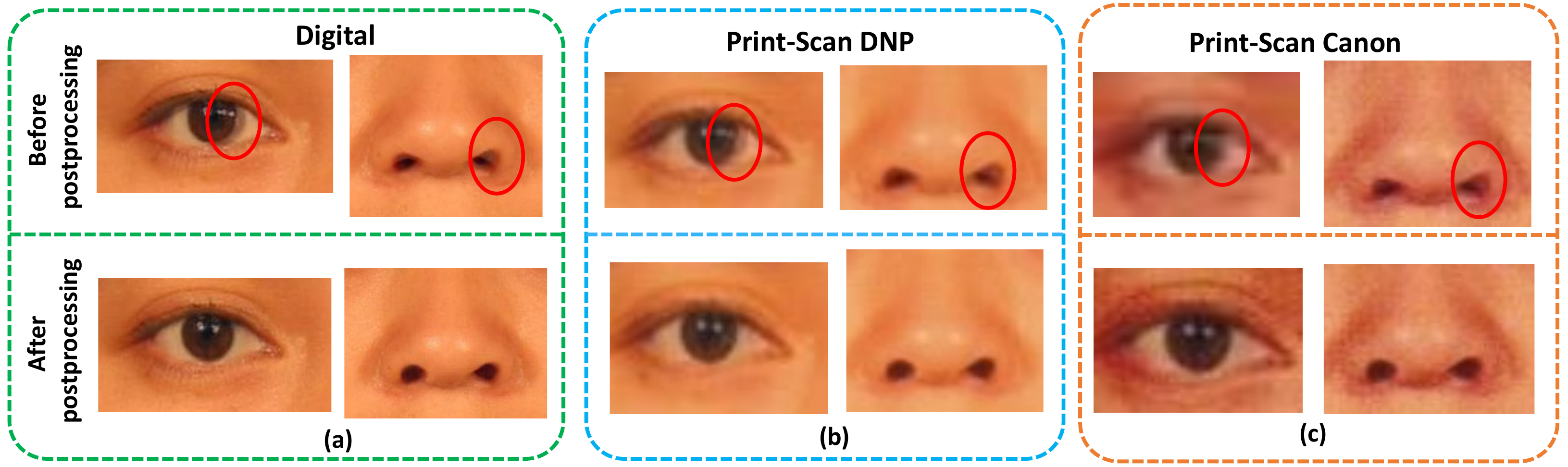}
\vspace{-3mm}
\caption{Illustration of issues of morphing before and after post-processing database from (i) Digital (ii) Print-Scan from DNP (PS-1) (iii) Print-Scan from Canon (PS-2)}
	\label{fig:morphissues}
\end{figure}

\begin{figure}[htp]
	\centering
	\includegraphics[width=1\linewidth]{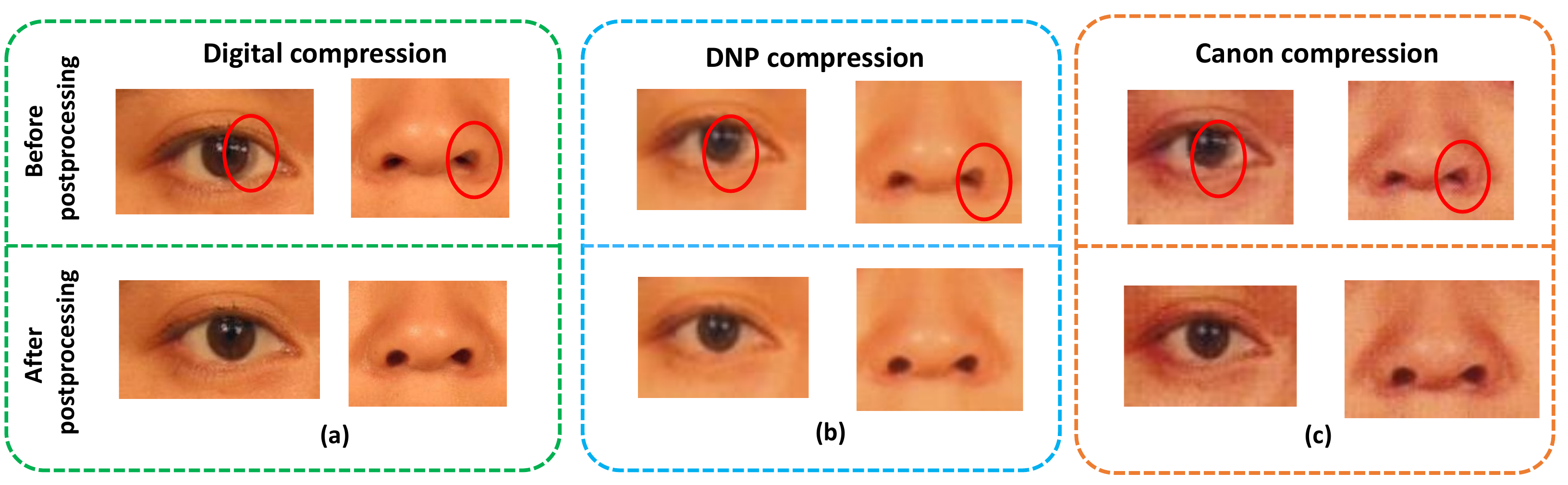}
\vspace{-3mm}
\caption{Illustration of before and after post-processing database from (i) Digital compression (ii) DNP compression (PS-1) (iii) Canon compression (PS-2)}
	\label{fig:compressionissues}
\end{figure}

\begin{itemize}
    
    \item \textbf{Q1} Does the performance of the proposed method improves when the morph attack detection is performed on post-processed morphing images compared with the morph images before post-processing?

    \item \textbf{Q2} Is the proposed method generalizable for morphed facial images generated from various mediums and the morphing images before and after post-processing? 
    
\end{itemize}

In the course of answering the research questions as mentioned above, the following are the main contributions of this work:
\begin{itemize}
 \item We present a novel S-MAD approach based on the multi-level fusion of ensemble features and classifiers to detect face-morphing attacks reliably. 
 \item We introduce a new dataset with manual post-processing to achieve high-quality face morphing images free from morphing noise and artefacts. The new dataset is collected using three different mediums that include both digital and two different printers. 
 \item Extensive experiments are carried out to benchmark the detection performance of the proposed method on three different mediums with and without post-processing. Further, the influence of image compression on detection performance is also benchmarked.
 \item The detection performance of the proposed method is benchmarked with the existing S-MAD techniques in two different experimental protocols.    
\end{itemize}

The rest of the paper is organised as follows: Section \ref{sec:db} details the newly generated dataset. Section \ref{sec:Pro} presents the proposed method using an ensemble of features and classifiers. Section \ref{sec:exp} details the experimental protocols and corresponding results. Section \ref{sec:discussion} provides a discussion on the observation made from the experimental results. Finally, Section \ref{sec:conc} concludes the current work. 

\section{Face Morphing Dataset}
\label{sec:db}
This section presents a new facial morphing dataset constructed using high-quality face images sampled from FRGC V2. The facial images are carefully selected to meet the enrolment guidelines, including zero pose, no shading on the face region, and no occlusion. The new dataset comprises 147 unique data subjects, further divided into two independent groups for training and testing. The training partition consists of 77 unique data subjects and the testing partition consists of 70 unique data subjects. In the next step, we perform the face morphing operation separately on the training and testing set. In this work, we employ the open-source face morphing tools \cite{Ferrara-FaceImageAlterations-Springer-2016, Ferrara-TextureBlendingAndShapeWarpingInFaceMorphing-IEEE-BIOSIG-2019} based on landmarks. Further, we have used only two face images with equal weights to perform morphing based on the earlier studies \cite{venkatesh-ageing-IJCB-2020, Raja-SOTAMD-DatabaseEvaluationBenchmarking-TIFS-2020} that have indicated high vulnerability on FRS.

In general, the morphing process will result in various types of noises, especially in the eyes and nose region. These noises include double edges in the eye region and the spreading of edges in the nose region. Figure \ref{fig:morphissues} illustrates the noises resulting from the morphing process that can be attributed to the variation in the geometry of the faces used for morphing. Even though these morphing noises are not common but exist in most cases, as shown in Figure \ref{fig:morphissues}, the morphing noises can also be predominantly observed even after the print-scan process. However, the quality of the print-scan process can also affect the visibility of edge spreading, as shown in Figure \ref{fig:morphissues}(c). Further, as noticed from Figure \ref{fig:compressionissues}, even after the images are compressed to follow the guidelines of ICAO \cite{ICAO-9303-p1-2015, ICAO-9303-p9-2015}, the morphing noises are still visible in both digital and print-scan versions. Therefore, it is essential to post-process the morphing face image to weed out these noises so that the human observer cannot identify the morphing based on these noises.

\begin{table}[htp]
  \centering
  \resizebox{0.9\linewidth}{!}
  {
    \begin{tabular}{|c|p{5em}|c|p{5em}|c|p{4.5em}|c|p{3em}|}
    \hline
    \rowcolor[rgb]{.851,.851,.851}  Image Type & before post-processing & {after post-processing} & Total\\
    \hline
    Digital images &  1071 & 1071 & 2142 \\
    \hline
    Print \& Scan &  1071X2 (printers) & 1071 X 2 (printers) & 4284 \\
    \hline
    Print \& Scan compression &  1071X2 (printers) & 1071X2 (printers) & 4284 \\
    \hline
    Total &  5355 & 5355 & 10710 \\
    \hline
    \end{tabular}
    }
  \caption{Total number of morphing images before and after manual post-processing.}
  \label{tab:morphs}
\end{table}

\begin{table}[htbp]
  \centering
  \caption{Database statistics:  training and testing partitions}
    \resizebox{01\linewidth}{!}
    {
    \begin{tabular}{|p{8.135em}|c|c|c|c|c|c|}
    \hline
    \multirow{3}[6]{*}{\textbf{Data Partition}} & \multicolumn{6}{p{25.41em}|}{\textbf{Data Type}} \bigstrut\\
\cline{2-7}    \multicolumn{1}{|c|}{} & \multicolumn{2}{p{8.47em}|}{\textbf{Digital}} & \multicolumn{2}{p{8.47em}|}{\textbf{PS-1}} & \multicolumn{2}{p{8.47em}|}{\textbf{PS-2}} \bigstrut\\
\cline{2-7}    \multicolumn{1}{|c|}{} & \multicolumn{1}{p{4.235em}|}{\textbf{Bona fide}} & \multicolumn{1}{p{4.235em}|}{\textbf{Morph}} & \multicolumn{1}{p{4.235em}|}{\textbf{Bona fide}} & \multicolumn{1}{p{4.235em}|}{\textbf{Morph}} & \multicolumn{1}{p{4.235em}|}{\textbf{Bona fide}} & \multicolumn{1}{p{4.235em}|}{\textbf{Morph}} \bigstrut\\
    \hline
    Training & 689   & 517   & 689   & 517   & 689   & 517 \bigstrut\\
    \hline
    Testing & 583   & 554   & 583   & 554   & 583   & 554 \bigstrut\\
    \hline
    \end{tabular}%
    }
  \label{tab:TrainTest}%
\end{table}%

Table \ref{tab:morphs} tabulates the statistics of the newly developed face morphing dataset with morphing samples before and after manual post-processing. The manual post-processing is carried out using Adobe Photoshop \cite{Adobe-photoshop} to obtain professional-quality passport face images. Figure \ref{fig:morphissues} and \ref{fig:compressionissues} illustrates the manual post-processing images in which the morphing noises are corrected to achieve the highest quality of the morphed face images. In this work, face morphing uses the alpha value (or morphing factor) of $0.5$ by considering the highest vulnerability demonstrated in several earlier works \cite{venkatesh-ageing-IJCB-2020, Raja-SOTAMD-DatabaseEvaluationBenchmarking-TIFS-2020}. 

We first generate the face morphing images separately for the training and testing sets. In the next step, we used two different printers,  which are DNP and a Canon printer, to digitize the digital images by print \& scan. The DNP printer used in this work is the dye-sublimation photo printer that can generate the highest quality passport face images and is widely deployed in photo studios. In contrast, the CANON PIXMA printer is a conventional inkjet printer used for printing passport face images. We term the data generated using the DNP printer as PS-1 and CANON PIXMA printer as PS-2, respectively. {Figure \ref{fig:compressionissues}} illustrates the example images from the newly developed datasets before and after manual post-processing.

\subsection{Dataset partition: Train and Test}
To effectively evaluate the Morph Attack Detection (MAD) algorithms, the whole dataset is partitioned into two independent sets: training and testing.  The training set consists of 77 unique data subjects and the testing partition consists of 70 unique data subjects. The morphing images are generated by using the data subjects within each partition. Thus, the training set comprises 689 bona fide and 517 morph face images.  Table \ref{tab:TrainTest} indicates the statistics of training and testing independently for morphing samples before and after manual post-processing. 

\begin{figure*}[htp]
	\centering
	\includegraphics[width=0.85\linewidth]{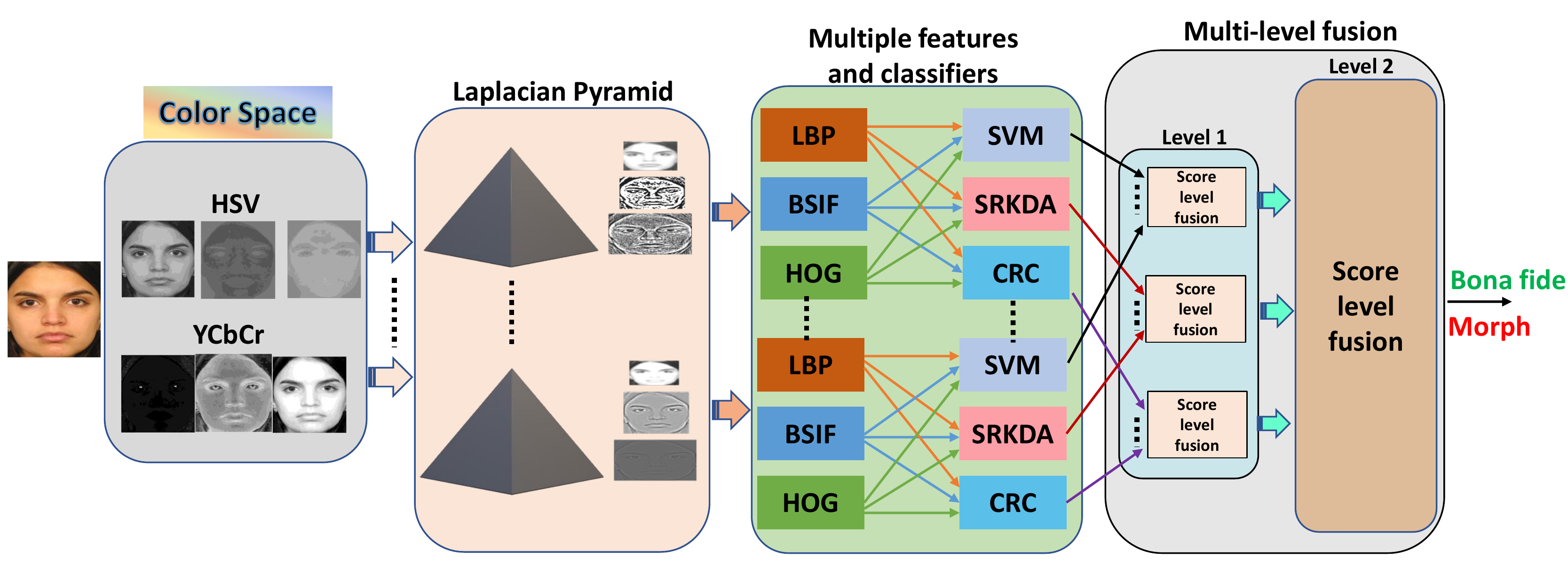}
\vspace{-3mm}
\caption{Block diagram of the proposed method}
	\label{fig:proposedmethod}
\end{figure*}
\section{Proposed Method}
\label{sec:Pro}
Figure \ref{fig:proposedmethod} shows the block diagram of the proposed method leveraged on the multi-level score level fusion of multiple features. The main objective of the proposed method is to exploit the complementary features of the different feature extractors and classifiers combined at two different levels. We assert that the use of complementary features and classification scores can capture the discriminant information useful for reliable face morph detection. The proposed method is designed using four different functional units, namely: (a) color space, (b) scale-space decomposition, (c) multiple features and classifiers (d) multi-level fusion. We discuss each of the functional units in detail in the following subsections. 
\subsection{Color space representation}
Given the input image $I$, the first step is to extract the different color spaces using $YC_{b}C_{r}$ and $HSV$.  We have selected these two color spaces by considering their robustness to capture the morphing noises as is demonstrated in earlier works \cite{Raghavendra-MADusingScaleSpaceFeature-ISBA-2019}. Thus, for the given image $I$, we get six different representations such as:$ I_{Col}$ = $I_{H}, I_{S}, I_{V}, I_{Y}, I_{C_{b}}, I_{C_{r}}$. 
\subsection{Scale-Space  decomposition}
In the next step, we extract the scale-space features on each color space image using the Laplacian pyramid \cite{LaplacianPyramid-1983}.  The choice of Laplacian pyramid-based scale-space features extraction is made by considering the effectiveness in extracting the discriminant features compared to similar techniques such as steerable pyramids \cite{Raghavendra-DetectingFaceMorphing-CVIP-2018}. We use three-level decomposition on each color image based on their empirical evaluation.  Thus, given the color image $ I_{H}$, the corresponding scale-space images can be represented as $ I_{H1}, I_{H2}, I_{H3}$. In this work, we have used six different color channels and thus, the corresponding scale-space representation will result in $6 \times 3 = 18$ sub-images that are independently processed to extract the multiple features. Let the sub-images be represented as: $SI_{k} = {SI_{1}, SI_{2,} \ldots, SI_{18}}, \forall k = {1,2, \ldots, 18}$.

\subsection{Multiple features and classifiers}
Multiple features and classification systems used in this work are based on three types of feature extraction and three different classifiers. Three different feature extraction techniques include Local Binary Patterns (LBP), Histogram of Gradients (HoG) and Binary Statistical Image Features (BSIF). These three features are selected by considering the complementary features that include texture features extracted using both hand-crafted and naturally learned in addition to the gradient information. These features represent the image's different characteristics, especially the pixel discontinuities, and thus can provide rich information to detect the morphing processing.  Given the sub-image $SI_{k}, \forall k = {1,2, \ldots, 18}$, three different types of features are extracted independently. 

In the next step, we employ three different types of classifiers, including linear Support Vector Machine (SVM) \cite{vapnik-svm-1999}, Spectral Regression Kernel Discriminant Analysis (SRKDA) \cite{cai2011speed} and Probabilistic Collaborative Representation Classifier (P-CRC) \cite{zhang-crc-2011}. We have considered these three classifiers by considering the high performance and robustness of various data sources{{\cite{Sushma_MAD_Survey}}. Further, the non-availability of the large-scale morphing database justifies the choice of the ensemble of these three classifiers to achieve reliable morph detection.  Given the features independently from the three different feature extraction techniques, we independently obtain the comparison scores from three different classification techniques. 
\subsection{Multi-level fusion}
This work proposes the two-level fusion of comparison scores obtained using multiple classifiers. The first level of fusion will combine the comparison scores obtained using individual classifiers corresponding to three different feature extraction techniques. Therefore, first-level fusion has three independent fusion units corresponding to three independent classifiers.  In the second level, we combine the comparison scores from the first level corresponding to individual classifiers to make the final decision. The multi-level fusion is designed based on empirical experiments that have indicated superior performance compared to serial fusion. At both levels, we have used the weighted sum rule to perform the fusion and weights are computed using the bootstrap method \cite{Raghavendra-PalmprintFusion-PR-2014} on the development dataset and kept constant through the experiments.

\section{Experiments and Results}
\label{sec:exp}

\begin{table}[htp]
  \centering
  \caption{Experiment-1: Quantitative results of MAD algorithms on different datasets}
  \resizebox{0.99\linewidth}{!}
    {
    \begin{tabular}{|c|c|c|p{7.935em}|c|c|c|c|c|c|}
    \hline
    \multicolumn{1}{|c|}{\multirow{4}[8]{*}{\textbf{Dataset}}} & \multicolumn{2}{c|}{\multirow{3}[6]{*}{\textbf{Post-processing}}} & \multirow{4}[8]{*}{\textbf{MAD Algorithms}} & \multicolumn{3}{p{12.705em}|}{\textbf{Detection Performance}} & \multicolumn{3}{p{12.705em}|}{\textbf{Detection Performance }} \bigstrut\\
\cline{5-10}          & \multicolumn{2}{c|}{} & \multicolumn{1}{c|}{} & \multicolumn{1}{r|}{\multirow{2}[4]{*}{\textbf{D-EER (\%)}}} & \multicolumn{2}{p{8.47em}|}{\textbf{BPCER @APCER }} & \multicolumn{1}{p{4.235em}|}{\textbf{D-EER (\%)}} & \multicolumn{2}{p{8.47em}|}{\textbf{BPCER @APCER  }} \bigstrut\\
\cline{6-10}          & \multicolumn{2}{c|}{} & \multicolumn{1}{c|}{} &       & \textbf{=5\%} & \textbf{=10\%} &       & \textbf{=5\%} & \textbf{=10\%} \bigstrut\\
\cline{2-3}\cline{5-10}          & \multicolumn{1}{p{4.235em}|}{\textbf{Training}} & \multicolumn{1}{p{4.235em}|}{\textbf{Testing}} & \multicolumn{1}{c|}{} & \multicolumn{3}{p{12.705em}|}{\textbf{without compression}} & \multicolumn{3}{p{12.705em}|}{\textbf{with compression}} \bigstrut\\  
    \hline
    \multicolumn{1}{|c|}{\multirow{6}[12]{*}{\textbf{Digital}}} & \multicolumn{1}{c|}{\multirow{3}[6]{*}{Before}} & \multicolumn{1}{c|}{\multirow{3}[6]{*}{Before}} & \textbf{Proposed Method} & \textbf{0} & \textbf{0} & \textbf{0} & \textbf{0} & \textbf{0} & \textbf{0} \bigstrut\\
\cline{4-10}          &       &       & Ensemble Features \cite{Venkatesh-SingleMAD-Fusion-2020} & 0.18  & 0     & 0     & 0.18  & 0     & 0 \bigstrut\\
\cline{4-10}          &       &       & Hybrid Features \cite{Raghavendra-MADusingScaleSpaceFeature-ISBA-2019} & 0     & 0     & 0     & 0.18  & 0     & 0 \bigstrut\\
\cline{2-10}          & \multicolumn{1}{c|}{\multirow{3}[6]{*}{After}} & \multicolumn{1}{c|}{\multirow{3}[6]{*}{After}} & \textbf{Proposed Method} & \textbf{0.18} & \textbf{0} & \textbf{0} & \textbf{0.36} & \textbf{0} & \textbf{0} \bigstrut\\
\cline{4-10}          &       &       & Ensemble Features \cite{Venkatesh-SingleMAD-Fusion-2020} & 0.18  & 0     & 0     & 0.36  & 0     & 0 \bigstrut\\
\cline{4-10}          &       &       & Hybrid Features \cite{Raghavendra-MADusingScaleSpaceFeature-ISBA-2019} & 0.18  & 0     & 0     & 0.18  & 0     & 0 \bigstrut\\
    \hline
    \multicolumn{1}{|c|}{\multirow{6}[12]{*}{\textbf{PS-1}}} & \multicolumn{1}{c|}{\multirow{3}[6]{*}{Before}} & \multicolumn{1}{c|}{\multirow{3}[6]{*}{Before}} & \textbf{Proposed Method} & \textbf{0} & \textbf{0} & \textbf{0} & \textbf{3.45} & \textbf{2.47} & \textbf{1.02} \bigstrut\\
\cline{4-10}          &       &       & Ensemble Features \cite{Venkatesh-SingleMAD-Fusion-2020} & 0     & 0     & 0     & 4.27  & 3.6   & 1.71 \bigstrut\\
\cline{4-10}          &       &       & Hybrid Features \cite{Raghavendra-MADusingScaleSpaceFeature-ISBA-2019} & 0     & 0     & 0     & 5     & 5.14  & 2.4 \bigstrut\\
\cline{2-10}          & \multicolumn{1}{c|}{\multirow{3}[6]{*}{After}} & \multicolumn{1}{c|}{\multirow{3}[6]{*}{After}} & \textbf{Proposed Method} & \textbf{0} & \textbf{0} & \textbf{0} & \textbf{3.09} & \textbf{2.22} & \textbf{1.02} \bigstrut\\
\cline{4-10}          &       &       & Ensemble Features \cite{Venkatesh-SingleMAD-Fusion-2020} & 0     & 0     & 0     & 3.28  & 2.4   & 1.54 \bigstrut\\
\cline{4-10}          &       &       & Hybrid Features \cite{Raghavendra-MADusingScaleSpaceFeature-ISBA-2019} & 0     & 0     & 0     & 4.46  & 4.28  & 2.74 \bigstrut\\
    \hline
    \multicolumn{1}{|c|}{\multirow{6}[12]{*}{\textbf{PS-2}}} & \multicolumn{1}{c|}{\multirow{3}[6]{*}{Before}} & \multicolumn{1}{c|}{\multirow{3}[6]{*}{Before}} & \textbf{Proposed Method} & \textbf{10.00} & \textbf{15.01} & \textbf{10.66} & \textbf{7.72} & \textbf{11.83} & \textbf{7.2} \bigstrut\\
\cline{4-10}          &       &       & Ensemble Features \cite{Venkatesh-SingleMAD-Fusion-2020} & 11.00   & 16.98 & 11.66 & 8.72  & 11.66 & 8.06 \bigstrut\\
\cline{4-10}          &       &       & Hybrid Features \cite{Raghavendra-MADusingScaleSpaceFeature-ISBA-2019} & 14.09 & 29.33 & 19.38 & 8.54  & 14.4  & 6.86 \bigstrut\\
\cline{2-10}          & \multicolumn{1}{c|}{\multirow{3}[6]{*}{After}} & \multicolumn{1}{c|}{\multirow{3}[6]{*}{After}} & \textbf{Proposed Method} & \textbf{5.74}  & \textbf{6.34} & \textbf{3.75}     & \textbf{5.19} & \textbf{5.14} & \textbf{2.91} \bigstrut\\
\cline{4-10}          &       &       & Ensemble Features \cite{Venkatesh-SingleMAD-Fusion-2020} & 6.01  & 7.03  & 4.11  & 5.19  & 5.14  & 3.77 \bigstrut\\
\cline{4-10}          &       &       & Hybrid Features \cite{Raghavendra-MADusingScaleSpaceFeature-ISBA-2019} & 8.56  & 12.34 & 7.2   & 5.64  & 6.17  & 2.91 \bigstrut\\
    \hline
    \end{tabular}%
    }
  \label{tab:intra}%
\end{table}%

\begin{table}[htbp]
  \centering
  \caption{Experiment-2: Quantitative performance of MAD algorithms on before post-processing data generated using different morphing types}
   \resizebox{0.8\linewidth}{!}
    {
    \begin{tabular}{|c|c|p{8.3em}|c|c|c|c|c|c|}
    \hline
    \multicolumn{1}{|c|}{\multirow{4}[8]{*}{\textbf{Training data}}} & \multicolumn{1}{c|}{\multirow{4}[8]{*}{\textbf{Testing Data}}} & \multirow{4}[8]{*}{\textbf{MAD Algorithms}} & \multicolumn{3}{p{12.705em}|}{\textbf{Detection Performance}} & \multicolumn{3}{p{12.705em}|}{\textbf{Detection Performance}} \bigstrut\\
\cline{4-9}          &       & \multicolumn{1}{c|}{} & \multicolumn{1}{r|}{\multirow{2}[4]{*}{\textbf{D-EER (\%)}}} & \multicolumn{2}{p{8.47em}|}{\textbf{BPCER @APCER  }} & \multicolumn{1}{p{4.235em}|}{\textbf{D-EER (\%)}} & \multicolumn{2}{p{8.47em}|}{\textbf{BPCER @APCER  }} \bigstrut\\
\cline{5-9}          &       & \multicolumn{1}{c|}{} &       & \textbf{=5\%} & \textbf{=10\%} &       & \textbf{=5\%} & \textbf{=10\%} \bigstrut\\
\cline{4-9}          &       & \multicolumn{1}{c|}{} & \multicolumn{3}{p{12.705em}|}{\textbf{without compression}} & \multicolumn{3}{p{12.705em}|}{\textbf{with compression}} \bigstrut\\

    \hline
    \multicolumn{1}{|c|}{\multirow{6}[12]{*}{\textbf{Digital}}} & \multicolumn{1}{c|}{\multirow{3}[6]{*}{PS-1}} & \textbf{Proposed Method} & \textbf{31.90} & \textbf{78.38} & \textbf{67.12} & \textbf{31.64} & \textbf{77.53} & \textbf{67.58} \bigstrut\\
\cline{3-9}          &       & Ensemble Features \cite{Venkatesh-SingleMAD-Fusion-2020} & 38.27 & 87.82 & 80.27 & 38.09 & 89.02 & 81.3 \bigstrut\\
\cline{3-9}          &       & Hybrid Features \cite{Raghavendra-MADusingScaleSpaceFeature-ISBA-2019} & 37.72 & 88.67 & 78.9  & 35.73 & 86.96 & 76.67 \bigstrut\\
\cline{2-9}          & \multicolumn{1}{c|}{\multirow{3}[6]{*}{PS-2}} & \textbf{Proposed Method} & \textbf{47.08} & \textbf{94.68} & \textbf{88.16} & \textbf{45.45} & \textbf{92.79} & \textbf{86.96} \bigstrut\\
\cline{3-9}          &       & Ensemble Features \cite{Venkatesh-SingleMAD-Fusion-2020} & 50 & 97.77 & 93.31 & 50 & 97.42 & 92.28 \bigstrut\\
\cline{3-9}          &       & Hybrid Features \cite{Raghavendra-MADusingScaleSpaceFeature-ISBA-2019} & 50 & 96.22 & 93.65 & 50    & 94.85 & 90.73 \bigstrut\\
    \hline
    \multicolumn{1}{|c|}{\multirow{6}[12]{*}{\textbf{PS-1}}} & \multicolumn{1}{c|}{\multirow{3}[6]{*}{Digital}} & \textbf{Proposed Method} & \textbf{3.63} & \textbf{2.22} & \textbf{0.68} & \textbf{5.26} & \textbf{5.48} & \textbf{3.77} \bigstrut\\
\cline{3-9}          &       & Ensemble Features \cite{Venkatesh-SingleMAD-Fusion-2020} & 8.09  & 14.23 & 6.51  & 8.09  & 15.6  & 6.51 \bigstrut\\
\cline{3-9}          &       & Hybrid Features \cite{Raghavendra-MADusingScaleSpaceFeature-ISBA-2019} & 8.54  & 13.2  & 7.54  & 20.72 & 43.91 & 33.1 \bigstrut\\
\cline{2-9}          & \multicolumn{1}{c|}{\multirow{3}[6]{*}{PS-2}} & \textbf{Proposed Method} & \textbf{20.9}  & \textbf{55.74} & \textbf{41.16} & \textbf{12.18} & \textbf{23.67} & \textbf{15.95} \bigstrut\\
\cline{3-9}          &       & Ensemble Features \cite{Venkatesh-SingleMAD-Fusion-2020} & 19.72 & 45.11 & 33.44 & 13.18 & 21.09 & 15.26 \bigstrut\\
\cline{3-9}          &       & Hybrid Features \cite{Raghavendra-MADusingScaleSpaceFeature-ISBA-2019} & 24.91 & 57.11 & 45.45 & 13.36 & 22.98 & 18.01 \bigstrut\\
    \hline
    \multicolumn{1}{|c|}{\multirow{6}[12]{*}{\textbf{PS-2}}} & \multicolumn{1}{c|}{\multirow{3}[6]{*}{Digital}} & \textbf{Proposed Method} & \textbf{9.45} & \textbf{16.63} & \textbf{8.74}  & \textbf{9.63} & \textbf{16.98} & \textbf{9.6} \bigstrut\\
\cline{3-9}          &       & Ensemble Features \cite{Venkatesh-SingleMAD-Fusion-2020} & 21.09 & 46.31 & 36.02 & 17.63 & 40.13 & 26.92 \bigstrut\\
\cline{3-9}          &       & Hybrid Features \cite{Raghavendra-MADusingScaleSpaceFeature-ISBA-2019} & 9.63  & 20.41 & 9.26  & 23.18 & 43.05 & 33.1 \bigstrut\\
\cline{2-9}          & \multicolumn{1}{c|}{\multirow{3}[6]{*}{PS-1}} & \textbf{Proposed Method} & \textbf{12.27} & \textbf{30.1}  & \textbf{18.09} & \textbf{0.16} & \textbf{0} & \textbf{0} \bigstrut\\
\cline{3-9}          &       & Ensemble Features \cite{Venkatesh-SingleMAD-Fusion-2020} & 14.27 & 28.64 & 19.72 & 0.16  & 0     & 0 \bigstrut\\
\cline{3-9}          &       & Hybrid Features \cite{Raghavendra-MADusingScaleSpaceFeature-ISBA-2019} & 19.72 & 42.19 & 31.73 & 0.72  & 0     & 0 \bigstrut\\
    \hline
    \end{tabular}%
    }
  \label{tab:onlybefore}%
\end{table}%

\begin{table}[htbp]
  \centering
  \caption{Experiment-2: Quantitative performance of MAD algorithms on after post-processing data generated using different morphing types}
   \resizebox{0.8\linewidth}{!}
    {
    \begin{tabular}{|c|c|p{8.365em}|c|c|c|c|c|c|}
    \hline
    \multicolumn{1}{|c|}{\multirow{4}[8]{*}{\textbf{Training data}}} & \multicolumn{1}{c|}{\multirow{4}[8]{*}{\textbf{Testing Data}}} & \multirow{4}[8]{*}{\textbf{MAD Algorithms}} & \multicolumn{3}{p{12.705em}|}{\textbf{Detection Performance}} & \multicolumn{3}{p{12.705em}|}{\textbf{Detection Performance }} \bigstrut\\
\cline{4-9}          &       & \multicolumn{1}{c|}{} & \multicolumn{1}{r|}{\multirow{2}[4]{*}{\textbf{D-EER (\%)}}} & \multicolumn{2}{p{8.47em}|}{\textbf{BPCER @APCER }} & \multicolumn{1}{p{4.235em}|}{\textbf{D-EER (\%)}} & \multicolumn{2}{p{8.47em}|}{\textbf{BPCER @APCER  }} \bigstrut\\
\cline{5-9}          &       & \multicolumn{1}{c|}{} &       & \textbf{=5\%} & \textbf{=10\%} &       & \textbf{=5\%} & \textbf{=10\%} \bigstrut\\
\cline{4-9}          &       & \multicolumn{1}{c|}{} & \multicolumn{3}{p{12.705em}|}{\textbf{without compression}} & \multicolumn{3}{p{12.705em}|}{\textbf{with compression}} \bigstrut\\
    \hline
    \multicolumn{1}{|c|}{\multirow{6}[12]{*}{\textbf{Digital}}} & \multicolumn{1}{c|}{\multirow{3}[6]{*}{PS-1}} & \textbf{Proposed Method} & \textbf{31.24} & \textbf{78.55} & \textbf{65.69} & \textbf{31.87} & \textbf{78.9} & \textbf{68.09} \bigstrut\\
\cline{3-9}          &       & Ensemble Features \cite{Venkatesh-SingleMAD-Fusion-2020} & 34.15 & 83.53 & 76.67 & 37.52 & 87.13 & 77.53 \bigstrut\\
\cline{3-9}          &       & Hybrid Features \cite{Raghavendra-MADusingScaleSpaceFeature-ISBA-2019} & 38.06 & 89.87 & 80.78 & 37.7  & 88.67 & 81.3 \bigstrut\\
\cline{2-9}          & \multicolumn{1}{c|}{\multirow{3}[6]{*}{PS-2}} & \textbf{Proposed Method} & \textbf{35.6} & \textbf{87.3} & \textbf{71.18} & \textbf{36.15} & \textbf{87.47} & \textbf{74.09} \bigstrut\\
\cline{3-9}          &       & Ensemble Features \cite{Venkatesh-SingleMAD-Fusion-2020} & 39.25 & 93.31 & 84.21 & 42.26 & 94.51 & 84.56 \bigstrut\\
\cline{3-9}          &       & Hybrid Features \cite{Raghavendra-MADusingScaleSpaceFeature-ISBA-2019} & 44.44 & 91.25 & 82.16 & 41.71 & 91.59 & 80.61 \bigstrut\\
    \hline
    \multicolumn{1}{|c|}{\multirow{6}[12]{*}{\textbf{PS-1}}} & \multicolumn{1}{c|}{\multirow{3}[6]{*}{Digital}} & \textbf{Proposed Method} & \textbf{4.09} & \textbf{3.94} & \textbf{2.91} & \textbf{6.82} & \textbf{9.43} & \textbf{5.83} \bigstrut\\
\cline{3-9}          &       & Ensemble Features \cite{Venkatesh-SingleMAD-Fusion-2020} & 9.37  & 15.43 & 8.06  & 8.19  & 12    & 12.17 \bigstrut\\
\cline{3-9}          &       & Hybrid Features \cite{Raghavendra-MADusingScaleSpaceFeature-ISBA-2019} & 20.03 & 32.76 & 27.44 & 28.14 & 60.2  & 49.05 \bigstrut\\
\cline{2-9}          & \multicolumn{1}{c|}{\multirow{3}[6]{*}{PS-2}} & \textbf{Proposed Method} & \textbf{12.12} & \textbf{25.27} & \textbf{15.32} & \textbf{7.47}  & \textbf{10.69} & \textbf{6.83} \bigstrut\\
\cline{3-9}          &       & Ensemble Features \cite{Venkatesh-SingleMAD-Fusion-2020} & 13.02 & 26.75 & 16.63 & 8.19  & 12    & 6.86 \bigstrut\\
\cline{3-9}          &       & Hybrid Features \cite{Raghavendra-MADusingScaleSpaceFeature-ISBA-2019} & 24.86 & 50.08 & 42.19 & 10.47 & 16.46 & 10.46 \bigstrut\\
    \hline
    \multicolumn{1}{|c|}{\multirow{6}[12]{*}{\textbf{PS-2}}} & \multicolumn{1}{c|}{\multirow{3}[6]{*}{Digital}} & \textbf{Proposed Method} & \textbf{10.47} & \textbf{21.09} & \textbf{11.49} & \textbf{11.84} & \textbf{23.15} & \textbf{13.2} \bigstrut\\
\cline{3-9}          &       & Ensemble Features \cite{Venkatesh-SingleMAD-Fusion-2020} & 18.48 & 45.11 & 31.9  & 15.93 & 38.59 & 25.38 \bigstrut\\
\cline{3-9}          &       & Hybrid Features \cite{Raghavendra-MADusingScaleSpaceFeature-ISBA-2019} & 11.1  & 26.92 & 13.89 & 23.13 & 44.94 & 36.87 \bigstrut\\
\cline{2-9}          & \multicolumn{1}{c|}{\multirow{3}[6]{*}{PS-1}} & \textbf{Proposed Method} & 13.93 & 28.98 & 19.55 & \textbf{0} & \textbf{0} & \textbf{0} \bigstrut\\
\cline{3-9}          &       & Ensemble Features \cite{Venkatesh-SingleMAD-Fusion-2020} & 9.92  & 18.18 & 9.94  & 0.16  & 0     & 0 \bigstrut\\
\cline{3-9}          &       & Hybrid Features \cite{Raghavendra-MADusingScaleSpaceFeature-ISBA-2019} & 18.48 & 41.16 & 30.7  & 0.55  & 0     & 0 \bigstrut\\
    \hline
    \end{tabular}%
    }
  \label{tab:onlyafter}%
\end{table}%

\begin{table}[htbp]
  \centering
  \caption{Experiment-3: Quantitative performance of MAD algorithms by training after post-processing data and testing before post-processing data generated using different morphing types}
  \resizebox{0.8\linewidth}{!}
    {
    \begin{tabular}{|c|c|p{8.235em}|c|c|c|c|c|c|}
    \hline
    \multicolumn{1}{|c|}{\multirow{4}[8]{*}{\textbf{Training data}}} & \multicolumn{1}{c|}{\multirow{4}[8]{*}{\textbf{Testing Data}}} & \multirow{4}[8]{*}{\textbf{MAD Algorithms}} & \multicolumn{3}{p{12.705em}|}{\textbf{Detection Performance}} & \multicolumn{3}{p{12.705em}|}{\textbf{Detection Performance }} \bigstrut\\
\cline{4-9}          &       & \multicolumn{1}{c|}{} & \multicolumn{1}{r|}{\multirow{2}[4]{*}{\textbf{D-EER (\%)}}} & \multicolumn{2}{p{8.47em}|}{\textbf{BPCER @APCER }} & \multicolumn{1}{p{4.235em}|}{\textbf{D-EER (\%)}} & \multicolumn{2}{p{8.47em}|}{\textbf{BPCER @APCER }} \bigstrut\\
\cline{5-9}          &       & \multicolumn{1}{c|}{} &       & \textbf{=5\%} & \textbf{=10\%} &       & \textbf{=5\%} & \textbf{=10\%} \bigstrut\\
\cline{4-9}          &       & \multicolumn{1}{c|}{} & \multicolumn{3}{p{12.705em}|}{\textbf{without compression}} & \multicolumn{3}{p{12.705em}|}{\textbf{with compression}} \bigstrut\\
    \hline
    \multicolumn{1}{|c|}{\multirow{6}[12]{*}{\textbf{Digital}}} & \multicolumn{1}{c|}{\multirow{3}[6]{*}{PS-1}} & \textbf{Proposed Method} & \textbf{30.72} & \textbf{80.44} & \textbf{68.95} & \textbf{31.9} & \textbf{80.44} & \textbf{67.92} \bigstrut\\
\cline{3-9}          &       & Ensemble Features \cite{Venkatesh-SingleMAD-Fusion-2020} & 36.18 & 86.96 & 79.07 & 37.72 & 88.67 & 79.41 \bigstrut\\
\cline{3-9}          &       & Hybrid Features \cite{Raghavendra-MADusingScaleSpaceFeature-ISBA-2019} & 37.54 & 89.7  & 81.3  & 35.99 & 86.1  & 78.55 \bigstrut\\
\cline{2-9}          & \multicolumn{1}{c|}{\multirow{3}[6]{*}{PS-2}} & \textbf{Proposed Method} & \textbf{46.45} & \textbf{94.16} & \textbf{88.67} & \textbf{46.63} & \textbf{93.31} & \textbf{87.99} \bigstrut\\
\cline{3-9}          &       & Ensemble Features \cite{Venkatesh-SingleMAD-Fusion-2020} & 50.27 & 97.25 & 93.31 & 51.27 & 96.91 & 92.1 \bigstrut\\
\cline{3-9}          &       & Hybrid Features \cite{Raghavendra-MADusingScaleSpaceFeature-ISBA-2019} & 50.9  & 94.16 & 90.39 & 52.09 & 95.54 & 91.76 \bigstrut\\
    \hline
    \multicolumn{1}{|c|}{\multirow{6}[12]{*}{\textbf{PS-1}}} & \multicolumn{1}{c|}{\multirow{3}[6]{*}{Digital}} & \textbf{Proposed Method} & \textbf{5.45} & \textbf{5.83} & \textbf{3.6} & \textbf{8.9} & \textbf{13.55} & \textbf{8.57} \bigstrut\\
\cline{3-9}          &       & Ensemble Features \cite{Venkatesh-SingleMAD-Fusion-2020} & 13.1  & 25.55 & 18.52 & 13.18 & 25.72 & 16.12 \bigstrut\\
\cline{3-9}          &       & Hybrid Features \cite{Raghavendra-MADusingScaleSpaceFeature-ISBA-2019} & 9.82  & 19.72 & 9.6   & 27.09 & 60.89 & 51.11 \bigstrut\\
\cline{2-9}          & \multicolumn{1}{c|}{\multirow{3}[6]{*}{PS-2}} & \textbf{Proposed Method} & \textbf{17.9}  & \textbf{37.77} & \textbf{26.7}  & \textbf{11.81} & \textbf{22.29} & \textbf{12.52} \bigstrut\\
\cline{3-9}          &       & Ensemble Features \cite{Venkatesh-SingleMAD-Fusion-2020} & 18.54 & 41.16 & 29.5  & 11.99 & 20.06 & 12.69 \bigstrut\\
\cline{3-9}          &       & Hybrid Features \cite{Raghavendra-MADusingScaleSpaceFeature-ISBA-2019} & 28.08 & 56.43 & 45.62 & 12.36 & 21.95 & 15.09 \bigstrut\\
    \hline
    \multicolumn{1}{|c|}{\multirow{6}[12]{*}{\textbf{PS-2}}} & \multicolumn{1}{c|}{\multirow{3}[6]{*}{Digital}} & \textbf{Proposed Method} & \textbf{11.81} & \textbf{21.44} & \textbf{14.92} & \textbf{13.72} & \textbf{31.73} & \textbf{21.09} \bigstrut\\
\cline{3-9}          &       & Ensemble Features \cite{Venkatesh-SingleMAD-Fusion-2020} & 22.08 & 50.94 & 39.1  & 18.81 & 44.59 & 28.47 \bigstrut\\
\cline{3-9}          &       & Hybrid Features \cite{Raghavendra-MADusingScaleSpaceFeature-ISBA-2019} & 13.36 & 29.15 & 17.15 & 22.63 & 46.68 & 34.47 \bigstrut\\
\cline{2-9}          & \multicolumn{1}{c|}{\multirow{3}[6]{*}{PS-1}} & \textbf{Proposed Method} & \textbf{12.72} & \textbf{21.56} & \textbf{13.32} & \textbf{0.3}  & \textbf{0}     & \textbf{0} \bigstrut\\
\cline{3-9}          &       & Ensemble Features \cite{Venkatesh-SingleMAD-Fusion-2020} & 13.18 & 22.81 & 16.46 & 0.16  & 0     & 0 \bigstrut\\
\cline{3-9}          &       & Hybrid Features \cite{Raghavendra-MADusingScaleSpaceFeature-ISBA-2019} & 18.99 & 40.13 & 30.7  & 0.72  & 0     & 0 \bigstrut\\
    \hline
    \end{tabular}%
    }
  \label{tab:taftertbef}%
\end{table}%

\begin{table}[htbp]
  \centering
  \caption{Experiment-3: Quantitative performance of MAD algorithms by training before post-processing data and testing after post-processing data generated using different morphing types}
   \resizebox{0.8\linewidth}{!}
    {
    \begin{tabular}{|c|c|p{8.065em}|c|c|c|c|c|c|}
    \hline
    \multicolumn{1}{|c|}{\multirow{4}[8]{*}{\textbf{Training data}}} & \multicolumn{1}{c|}{\multirow{4}[8]{*}{\textbf{Testing Data}}} & \multirow{4}[8]{*}{\textbf{MAD Algorithms}} & \multicolumn{3}{p{12.705em}|}{\textbf{Detection Performance}} & \multicolumn{3}{p{12.705em}|}{\textbf{Detection Performance }} \bigstrut\\
\cline{4-9}          &       & \multicolumn{1}{c|}{} & \multicolumn{1}{r|}{\multirow{2}[4]{*}{\textbf{D-EER (\%)}}} & \multicolumn{2}{p{8.47em}|}{\textbf{BPCER @APCER }} & \multicolumn{1}{p{4.235em}|}{\textbf{D-EER (\%)}} & \multicolumn{2}{p{8.47em}|}{\textbf{BPCER @APCER }} \bigstrut\\
\cline{5-9}          &       & \multicolumn{1}{c|}{} &       & \textbf{=5\%} & \textbf{=10\%} &       & \textbf{=5\%} & \textbf{=10\%} \bigstrut\\
\cline{4-9}          &       & \multicolumn{1}{c|}{} & \multicolumn{3}{p{12.705em}|}{\textbf{without compression}} & \multicolumn{3}{p{12.705em}|}{\textbf{with compression}} \bigstrut\\
    \hline
    \multicolumn{1}{|c|}{\multirow{6}[12]{*}{\textbf{Digital}}} & \multicolumn{1}{c|}{\multirow{3}[6]{*}{PS-1}} & \textbf{Proposed Method} & \textbf{32.05} & \textbf{79.93} & \textbf{69.12} & \textbf{32.6} & \textbf{79.07} & \textbf{71.01} \bigstrut\\
\cline{3-9}          &       & Ensemble Features \cite{Venkatesh-SingleMAD-Fusion-2020} & 36.52 & 87.82 & 80.96 & 38.88 & 89.36 & 80.96 \bigstrut\\
\cline{3-9}          &       & Hybrid Features \cite{Raghavendra-MADusingScaleSpaceFeature-ISBA-2019} & 39.25 & 89.87 & 83.87 & 37.7  & 88.5  & 80.78 \bigstrut\\
\cline{2-9}          & \multicolumn{1}{c|}{\multirow{3}[6]{*}{PS-2}} & \textbf{Proposed Method} & \textbf{37.25} & \textbf{90.05} & \textbf{75.64} & \textbf{36.15} & \textbf{88.67} & \textbf{74.95} \bigstrut\\
\cline{3-9}          &       & Ensemble Features \cite{Venkatesh-SingleMAD-Fusion-2020} & 40.61 & 95.54 & 87.13 & 42.34 & 95.71 & 86.44 \bigstrut\\
\cline{3-9}          &       & Hybrid Features \cite{Raghavendra-MADusingScaleSpaceFeature-ISBA-2019} & 45.44 & 93.13 & 83.87 & 41.16 & 91.93 & 80.96 \bigstrut\\
    \hline
    \multicolumn{1}{|c|}{\multirow{6}[12]{*}{\textbf{PS-1}}} & \multicolumn{1}{c|}{\multirow{3}[6]{*}{Digital}} & \textbf{Proposed Method} & \textbf{3.46} & \textbf{2.57} & \textbf{1.02} & \textbf{5.64} & \textbf{6.68} & \textbf{3.94} \bigstrut\\
\cline{3-9}          &       & Ensemble Features \cite{Venkatesh-SingleMAD-Fusion-2020} & 8.92  & 15.6  & 7.54  & 9.92  & 16.63 & 9.77 \bigstrut\\
\cline{3-9}          &       & Hybrid Features \cite{Raghavendra-MADusingScaleSpaceFeature-ISBA-2019} & 17.3  & 32.76 & 24.52 & 22.13 & 45.45 & 34.81 \bigstrut\\
\cline{2-9}          & \multicolumn{1}{c|}{\multirow{3}[6]{*}{PS-2}} & \textbf{Proposed Method} & \textbf{16.4}  & \textbf{40.96} & \textbf{29.81} & \textbf{10.29} & \textbf{20.92} & \textbf{10.97} \bigstrut\\
\cline{3-9}          &       & Ensemble Features \cite{Venkatesh-SingleMAD-Fusion-2020} & 17.66 & 42.02 & 31.73 & 10.65 & 18.01 & 11.32 \bigstrut\\
\cline{3-9}          &       & Hybrid Features \cite{Raghavendra-MADusingScaleSpaceFeature-ISBA-2019} & 24.31 & 50.08 & 41.16 & 11.84 & 19.03 & 13.55 \bigstrut\\
    \hline
    \multicolumn{1}{|c|}{\multirow{6}[12]{*}{\textbf{PS-2}}} & \multicolumn{1}{c|}{\multirow{3}[6]{*}{Digital}} & \textbf{Proposed Method} & \textbf{9.74} & \textbf{16.46} & \textbf{8.91} & \textbf{10.47} & \textbf{17.32} & \textbf{10.97} \bigstrut\\
\cline{3-9}          &       & Ensemble Features \cite{Venkatesh-SingleMAD-Fusion-2020} & 22.41 & 47.51 & 37.9  & 17.66 & 42.53 & 28.47 \bigstrut\\
\cline{3-9}          &       & Hybrid Features \cite{Raghavendra-MADusingScaleSpaceFeature-ISBA-2019} & 9.74  & 25.9  & 9.6   & 24.04 & 47.68 & 37.56 \bigstrut\\
\cline{2-9}          & \multicolumn{1}{c|}{\multirow{3}[6]{*}{PS-1}} & \textbf{Proposed Method} & \textbf{12.48} & \textbf{23.04} & \textbf{16.26} & \textbf{0.16} & \textbf{0} & \textbf{0} \bigstrut\\
\cline{3-9}          &       & Ensemble Features \cite{Venkatesh-SingleMAD-Fusion-2020} & 13.93 & 26.75 & 18.01 & 0.16  & 0     & 0 \bigstrut\\
\cline{3-9}          &       & Hybrid Features \cite{Raghavendra-MADusingScaleSpaceFeature-ISBA-2019} & 20.94 & 44.25 & 34.3  & 0.55  & 0     & 0 \bigstrut\\
    \hline
    \end{tabular}%
    }
  \label{tab:tbeftafter}%
\end{table}%
In this section, we present and discuss the proposed method's quantitative results and the existing methods such as Hybrid features \cite{Raghavendra-MADusingScaleSpaceFeature-ISBA-2019} and  Ensemble features \cite{Venkatesh-SingleMAD-Fusion-2020}. We particularly select these two existing methods as 
(1) these methods indicate the best performance in several reported studies \cite{Sushma_MAD_Survey}  and one of them is benchmarked on the  NIST FRVT morph \cite{MeiNGAN-morph-FRVT-2020} (2) these methods are based on the hand-crafted features thus are more appropriate to be compared with the proposed method (3) these methods are more appropriate by considering the size of the databases used in this work. The use of deep learning methods may result in overfitting due to the small datasets.  
The performance of the S-MAD techniques is benchmarked using  ISO/IEC 30107-3 \cite{ISO-IEC-30107-3-PAD-metrics-170227}  metrics such as  Attack Presentation Classification Error Rate (APCER (\%)), Bona fide Presentation Classification Error Rate (BPCER(\%)) and Detection-Equal Error Rate (D-EER(\%)).

\subsection{Experimental protocols:}
\label{sec:expPro}
To effectively evaluate the performance of the MAD algorithms using the proposed method, our experiments are categorized into three different protocols discussed as follows:  

\begin{itemize}
    \item \textbf{Experiment-1: Intra-dataset evaluation:} is performed within the same dataset type. This evaluation protocol performs training and testing on the same dataset type. As shown in Table \ref{tab:intra}, the three dataset types (digital, PS-I and PS-II) are independently evaluated before and after post-processing. For instance, the digital dataset type before post-processing is trained and the same dataset type is tested. A similar protocol is followed for the digital dataset type after post-processing, followed by the two different print-scan dataset types PS-I (before and after post-processing) and PS-II (before and after post-processing). All experiments are carried out with and without compression. 
\item \textbf{Experiment-2: Inter-medium evaluation}: is performed to analyze the MAD performance of the proposed method in cross-dataset types. This protocol is designed to investigate the robustness of the proposed method when it is trained and tested on different dataset types (digital, PS-1 and PS-2) generated from different mediums (digital, print-scan with and without compression). Tables \ref{tab:onlybefore} and \ref{tab:onlyafter} indicates the two different experiments performed for cross-dataset evaluation in the inter-medium scenario. Among the three dataset types employed in this work, we train one dataset type and test it on the other two. For instance, if the digital dataset type is trained, the two different print-scan dataset types, PS-I and PS-II, are tested. The same evaluation protocol is followed for the two print-scan dataset types. To better evaluate the cross-dataset performance of the proposed method, we have performed two different experiments (i) inter-medium evaluation before post-processing and (ii) inter-medium evaluation after post-processing.

    \item \textbf{Experiment-3: Inter-medium varied post-processing}: is performed to evaluate the performance of MAD in cross datasets generated from various mediums (digital, print-scan with and without compression) in both before and after post-processing scenarios. Tables \ref{tab:taftertbef} and \ref{tab:tbeftafter} indicates the two experiments conducted for inter-medium and varied post-processing scenario. Two different experiments were conducted to evaluate the proposed method's performance. Following the similar experimental protocol as inter-medium evaluation, the first experiment is performed by (i) training the dataset types after post-processing and testing the dataset types before post-processing. The second experiment is performed by training the dataset types before post-processing and testing after post-processing. 
    \end{itemize}
    
\subsection{Experimental results}
 \label{results}
 In this section, we present the quantitative results of the proposed method and the existing methods of the three different evaluation protocols.  The quantitative results obtained from the three different protocols designed for intra-dataset evaluation, inter-medium evaluation and inter-medium with varied post-processing evaluation scenarios are tabulated in the Tables \ref{tab:intra}, \ref{tab:onlybefore}, \ref{tab:onlyafter}, \ref{tab:taftertbef}, \ref{tab:tbeftafter}. 
   
\subsubsection{\textbf{Results on Experiment-1: Intra-dataset evaluation}}
   Based on the obtained results presented in Table \ref{tab:intra} following are the main observations: 
   
   \begin{itemize}
   
       \item The proposed method has indicated the best performance on all three data mediums before and after post-processing. Thus, the proposed method has emerged as the best-performing method before and after post-processing. 

            \item The detection performance of the existing methods also indicates the competitive performance, especially with digital and PS-1 data mediums both before and after post-processing. 
       
       \item The detection performance of the S-MAD techniques indicates the degraded performance, especially with the PS-2 data medium that can be noticed before and after post-processing data. Thus, the morph generation quality will impact the detection accuracy of both the proposed and existing S-MAD techniques.

       \item Performing the post-processing indicates the impact on the detection performance. In some cases, the detection performance of the proposed method and the existing methods indicates improvement. This can be attributed to the possible variations in the image quality that might have resulted from post-processing operation. However, with data compression, the performance difference is not noticeable. 
       
       \item The performance of the S-MAD algorithms also varies with and without compression, irrespective of the post-processing.
 \end{itemize} 
\subsubsection{\textbf{Results on Experiment-2: Inter-medium evaluation}}
Table \ref{tab:onlybefore} and \ref{tab:onlyafter} indicates the quantitative performance of the proposed method together with existing methods in Experiment 2. Based on the obtained results following can be noted: 

 \begin{itemize}
 
 \item The Inter-medium training and testing indicate the drastic degradation of the detection accuracy of both the proposed method and the existing methods. The degradation is noticed both before and after post-processing. 
 
 \item The S-MAD algorithms degrade more when algorithms are trained with digital and tested against PS-I and PS-II. Less degradation is noted when S-MAD algorithms are trained with PS-I and tested against digital and PS-II. Similar degradation is noticed both before and after post-processing.
 
 \item The S-MAD algorithms have indicated a better detection accuracy on the print-scan compression when compared to without compression, especially on the before post-processing data. However, the S-MAD algorithms did not show much difference in the detection performance on the before post-processing data. This indicates that using the post-processing data to train and test the S-MAD algorithms might be key to achieving the generalisation in cross-medium experiments.
 
 \item Based on the experimental results in Experiment-2, the proposed method has indicated the best performance compared to existing methods on both before and after post-processing data.  
 
 \end{itemize}

 \subsubsection{\textbf{Experiment-3: Inter-medium varied post-processing}}
In this section, we discuss the quantitative results of the proposed method and the existing S-MAD techniques, especially to study the influence of post-processing operation versus different mediums on detection accuracy. Tables \ref{tab:taftertbef}  and \ref{tab:tbeftafter} indicate the quantitative results of the S-MAD techniques, including the proposed method. Based on the obtained results, the following can be noted: 
 \begin{itemize}
\item The performance of the S-MAD algorithms indicates the degraded detection rate irrespective of the data post-processing type.  
\item In general, the performance of the S-MAD algorithms, including the proposed method, indicates the marginal improvement in the detection performance when trained using post-processed data irrespective of the data medium. 
\item The performance of the proposed method indicates the best performance compared with the existing methods, irrespective of the data type (before or after post-processing) used for the training. The best performance of the proposed method is when  PS-1 is trained and tested on digital data before and after post-processing.
 \end{itemize}

\section{Discussion}
 \label{sec:discussion}
The research questions formulated in Section \ref{sec:introduction} are answered below based on the extensive experiments conducted, obtained results and the observations made above.
\begin{itemize}
	\item {{\textbf{Q1}}. Does the performance of the proposed method improve when the morph attack detection is performed on post-processed morphed images when compared with the morph images before post-processing?}
	\begin{itemize}
		\item As noted by the obtained experimental results reported in Table  \ref{tab:onlybefore} \ref{tab:onlyafter}, the performance of the proposed method shows a marginal improvement when used with the morph images after post-processing in Experiment-1, especially on the PS-2 data medium. However, the proposed method's performance did not significantly influence (even though the proposed method has shown little improvement in some cases) the post-processing in Experiment-2. 
	\end{itemize}
	\item {\textbf{Q2}.Is the proposed method generalizable for morphed facial images generated from various mediums and also for the morphed images before and after post-processing?}
	\begin{itemize}
		\item Based on the experimental results (see Table \ref{tab:intra} \ref{tab:onlybefore}, \ref{tab:onlyafter}, \ref{tab:taftertbef}, \ref{tab:tbeftafter}), the proposed method has indicated the best performance in two different experimental protocols.  
	\end{itemize}     
\end{itemize}
{Thus, based on the obtained results, one can attribute the improvements to using multiple features with multiple classifiers, which would increase generalization.}

{\section{Conclusions and Future Work}}
\label{sec:conc}
Reliable face morphing attack detection using a single image is a challenging problem due to the variation in image quality attributed to the various source of the morph generation and digitisation processes. In this work, we proposed a new framework for S-MAD using multiple features and classifiers whose comparison scores are combined at multiple levels to detect face-morphing attacks reliably. We have also introduced a new dataset based on manual post-processing to generate high-quality face morphing images free from morphing artefacts. The dataset constructed has three different mediums: digital, Print-Scan (PS-1 re-digitised using DNP printer and PS-2 re-digitised using CANON printer) and print-scan compression. Extensive experiments are carried out using two different evaluation protocols to benchmark the performance of the proposed method together with the existing methods. The obtained results demonstrated the best performance of the proposed method in two different evaluation protocols compared with the existing methods.  {In future work, we could evaluate more advanced fusion techniques, benchmarking the proposed method and comparison with more SOTA approaches.}

\bibliographystyle{ieee}
\bibliography{FaceMorph}
\end{document}